\documentclass[conference,10pt]{IEEEtran} %
\IEEEoverridecommandlockouts
\usepackage{cite}
\usepackage[cmintegrals]{newtxmath}
\usepackage[T1]{fontenc} 
\usepackage{amsmath,amsfonts} 
\usepackage{graphicx}
\usepackage{textcomp}

\ifCLASSOPTIONcompsoc
\usepackage[caption=false,font=normalsize,labelfon
t=sf,textfont=sf]{subfig}
\else
\usepackage[caption=false,font=footnotesize]{subfi
g}
\fi

\usepackage[inline]{enumitem}
\usepackage{stackrel}
\usepackage{mathtools}
\usepackage[colorinlistoftodos]{todonotes}
\usepackage{clrscode3e}
\usepackage{multirow}

\usepackage{tikz}
\usetikzlibrary{calc,positioning}
\tikzstyle{vertex} = [draw, circle, minimum size=0.5cm,fill=lightgray!50!white]
\tikzstyle{label} = [text=blue, font=\scriptsize]
\tikzstyle{prob} = [text=black, font=\scriptsize]
\tikzstyle{action} = [text=red, font=\scriptsize]
\tikzstyle{bullet} = [fill, color=black, inner sep=0pt, minimum size=0.1cm, circle]

\newtheorem{lemma}{Lemma}[section]
\newtheorem{definition}{Definition}[section]

\newtheorem{remark}{Remark}[section]

\newcommand{\cf}{\emph{cf.}}

\DeclareMathOperator*{\argmax}{arg\,max}
\newcommand{\distr}[1]{\mathcal{D}(#1)}

\newcommand{\M}{\mathcal{M}} 
\newcommand{\N}{\mathcal{N}} 
\newcommand{\hyp}{\mathcal{H}} 
\newcommand{\err}{\ell_{err}}  
\newcommand{\mcbw}{\proc{Mc-BW}}
\newcommand{\mdpbw}{\proc{Mdp-BW}}

\newcommand{\T}{\mathcal{T}}
\newcommand{\R}{\mathcal{R}}
\newcommand{\LTL}{\mathcal{L}}
\newcommand{\bigo}{\mathcal{O}}

\newcommand{\Pb}{Pr}

\newcommand{\Obs}{\mathbf{Obs}}
\newcommand{\Path}{\mathbf{Paths}}

\newcommand{\X}{\mathcal{X}}

\begin{document}

\title{Active Learning of Markov Decision Processes using Baum-Welch algorithm (Extended)
\thanks{R.\ Reynouard and  A.\ Ing{\'{o}}lfsd{\'{o}}ttir have been supported by the project \emph{Learning and Applying Probabilistic Systems} (nr.\ 206574-051) of the Icelandic Research Fund. K.G. Larsen has been supported by the ERC Advanced Grant LASSO (nr.\ 669844), and the Innovation Fund Denmark center DiCyPS.}}

\author{\IEEEauthorblockN{Giovanni Bacci}
\IEEEauthorblockA{Dept. of Computer Science \\
Aalborg, Denmark \\ Email: {giovbacci@cs.aau.dk}}
\and
\IEEEauthorblockN{Anna Ing{\'{o}}lfsd{\'{o}}ttir}
\IEEEauthorblockA{Dept. of Computer Science \\ 
Reykjav\'{i}k, Iceland \\
Email: {annai@ru.is}}
\and
\IEEEauthorblockN{Kim G.\ Larsen}
\IEEEauthorblockA{Dept. of Computer Science \\
Aalborg, Denmark \\ Email: {kgl@cs.aau.dk}}
\and
\IEEEauthorblockN{Rapha\"{e}l Reynouard}
\IEEEauthorblockA{Dept. of Computer Science \\ 
Reykjav\'{i}k, Iceland \\
Email: {raphal20@ru.is}}
}


\maketitle

\begin{abstract}
Cyber-physical systems (CPSs) are naturally modelled as reactive systems with nondeterministic and probabilistic dynamics. Model-based verification techniques have proved effective in the deployment of safety-critical CPSs. 
Central for a successful application of such techniques is the construction of an accurate formal model for the system. Manual construction can be a resource-demanding and error-prone process, thus motivating the design of automata learning algorithms to synthesise a system model from observed system behaviours.

This paper revisits and adapts the classic Baum-Welch algorithm for learning Markov decision processes and Markov chains. For the case of MDPs, which typically demand more observations, we present a model-based active learning sampling strategy that choses examples which are most informative w.r.t.\ the current model hypothesis.  
We empirically compare our approach with state-of-the-art tools and demonstrate that the proposed active learning procedure can significantly reduce the number of observations required to obtain accurate models.
\end{abstract}


\section{Introduction}
Model-based verification techniques have proved effective in the deployment of safety-critical cyber-physical systems. Due to their interactions with a physical environment, CPSs are naturally modelled as reactive systems with nondeterministic and probabilistic dynamics. A popular formalism for such systems are discrete-time Markov decision processes (MDPs). 

Quantitative verification techniques like probabilistic model checking can provide strategies that are provably optimal with respect to the probability of satisfaction of some requirements expressed as LTL or PCTL formulae. Model checking tools such as \textsc{Prism}~\cite{KwiatkowskaNP11}, \textsc{Storm}~\cite{DehnertJK017}, 
and \textsc{Uppaal-Stratego}~\cite{DavidJLMT15} offer efficient methods for finite MDPs. These techniques assume that the model is an accurate formalisation of the true system. 
Thus, central for model-based verification is the construction of accurate models. 

Manual construction requires one to determine a big number of model parameters which can be a resource-demanding and error-prone process. This motivated the design of automata learning algorithms able to synthesise Markov chains~\cite{CarrascoO94,CarrascoO99} and deterministic Markov decision processes~\cite{MaoCJNLN11,ChenN12,MaoCJNLN16,TapplerA0EL19} from observed system behaviours.
These algorithms, in the large sample limit, identify the original (canonical) model. 
However, for practical applications, the available data is often limited, as the generation of a large number of observations can be a resource-demanding task. Additionally, there might be requirements on the size of the learned model, e.g., when the model has to be stored in an embedded system.

The Baum-Welch algorithm~\cite{Rabiner89} is an expectation maximisation technique~\cite{Dempster77} for learning model parameters of a hidden Markov model. This algorithm has recently been applied in model-based statistical verification of CPSs~\cite{KalajdzicJLBLSG16}, model checking of interval Markov chains~\cite{BenediktLW13}, and metric-based approximate minimisation of Markov chains~\cite{BacciBLM18}.

This paper proposes a variant of the Baum-Welch algorithm that learns model parameters for Markov chains and Markov decision processes from observed systems behaviours. 
As the original algorithm, it starts from a given model hypothesis and iteratively updates its transition probabilities until the likelihood of the data stops improving more than a suitably small $\epsilon$. 
The algorithm can be combined with other learning techniques like \textsc{Alergia}~\cite{CarrascoO94} and \textsc{IOAlergia}~\cite{MaoCJNLN11,ChenN12,MaoCJNLN16} for the choice of the initial hypothesis. 
Notably, by fixing a suitably small initial hypothesis, the algorithm can also be used to construct succinct, yet accurate, approximations of complex systems. This characteristic is particularly useful when one needs to control the size of the learned model e.g., to store it into an embedded system. 

Empirical comparisons with state-of-the-art tools show that the Baum-Welch algorithm for MDPs can achieve a better ratio of accuracy to the size of the model. However, when the size of initial hypothesis model is bigger than that of the system under learning it is not uncommon for the Baum-Welch algorithm to overfit the observation set. 

Learning MDPs typically requires more observations as the number of model parameters grows with the number of nondeterministic actions. To address this issue, we employ \emph{active learning}. Rather than collecting data samples at random, we steer the sampling of new observations aiming at uncovering unobserved behaviours, thus improving the accuracy of the current model hypothesis. In this line, we propose to learn an 
initial hypothesis from a relatively small set of system observations sampled at random. Then, for each hidden state we compute the expected number of times each action has been chosen from that state. This information is used to devise an observation-based scheduler aimed at 
restoring balance in the count of actions performed from each hidden state. This helps the collected data set to represent a wider spectrum of the nondeterministic behaviours of the systems under learning.

Experiments show that our active learning procedure can significantly reduce the number of observations required to obtain accurate models, achieving a faster convergence rate than that observed when employing uniform schedulers.

\paragraph*{Other Related Work} An influential active automata learning technique is Angluin's $L^*$-algorithm~\cite{Angluin87} for learning regular languages, which inspired a number of extensions better suited for modelling reactive systems~\cite{SteffenHM11,IsbernerHS14,CasselHJS16}. In this line of research, Tappler et al.~\cite{TapplerA0EL19} proposed an $L^{*}$-based technique for learning (deterministic) MDPs. The method iteratively refines the current hypothesis until the teacher cannot provide a counterexample sequence. 
For each refinement step a predefined amount of new observations is collected. In contrast to our proposal, new sequences are sampled targeting a subset of states that are marked as rare.

Other related work include \emph{model-based} learning techniques for partially observable MDPs (e.g.,\cite{ShaniBS05}). These techniques aim at learning how to act in an unknown partially observable domain taking actions based on an approximate model of the domain. Typically, they learn only a portion of the real model that is sufficient to optimise the strategy, leaving unnecessary parts of the system unexplored. In contrast, we aim at learning the whole model and be able to analyse it.


\section{Preliminaries and Notation}\label{sec:prelim}

We denote by $\mathbb{R}$, $\mathbb{Q}$, and $\mathbb{N}$ respectively the sets of real, rational, and natural numbers. 
We denote by $\Sigma^n$, $\Sigma^*$ and, $\Sigma^\omega$ respectively the set of words of length $n \in \mathbb{N}$, finite length, and infinite length, built over the finite alphabet $\Sigma$.


We denote by $\distr{\Omega}$ the set of discrete probability distributions on $\Omega$ 
For $x \in \Omega$, the \emph{Dirac distribution} concentrated at $x$ is the distribution $1_x \in \distr{\Omega}$ defined, for arbitrary $y \in \Omega$, as $1_x(y) = 1$ if $x = y$, $0$ otherwise.

\subsection{Markov decision processes and schedulers}

\begin{definition}
A \emph{discrete-time Markov decision process} is a tuple, $\M = \langle S, L, A, \iota, \{\tau_a\}_{a \in A} \rangle$, where
\begin{enumerate*}[label=(\roman*)]
  \item $S$ is a finite nonempty set of states,
  \item $L$ is a finite nonempty set of labels,
  \item $A$ is a finite nonempty set of actions,
  \item $\iota \in \distr{L \times S}$ is an initial distribution, and
  \item $\tau_a \colon S \rightarrow \distr{L \times S}$ is a probabilistic transition function.
\end{enumerate*}
\end{definition}

Intuitively, $\M$ initially emits a label and probabilistically moves to some state according to $\iota$. Then, if $\M$ is in state $s$ and receives an input action $a \in A$, it emits a label $\ell \in L$ and moves to state $s'$ with probability $\tau_a(s)(\ell, s')$. In this sense, $\M$ can be thought of as a state-machine that reacts to a stream of input actions $a_1, a_2, \cdots \in A^{\omega}$ by emitting \emph{traces} of labels of the form $\ell_1, \ell_2, \cdots \in L^\omega$.
\begin{remark}
We do not assume to know a priori which actions are available from a given state $s$ of the model. Rather, we assume the model to react with an error label, denoted $\err \in L$, and move back to $s$ with probability $1$ whenever an action $a \in A$ which is not available is chosen from the current state $s$. Formally, $a \notin \mathit{Available}(s)$ implies $\tau_a(s)(\err, s) =1$. 
\end{remark}

A path is an infinite sequence in $\Path = (L \times S \times A)^\omega$ representing an execution of $\M$.
We denote by $\Path_{\text{fin}} = (L \times S \times A)^*(L \times S)$ the set of finite paths. Analogously, we define the set of infinite (resp.\ finite) observations as $\Obs = (L \times A)^\omega$ (resp.\ $\Obs_{\text{fin}} = (L \times A)^*L$). 
The length of a finite path $w$ (resp.\ observation $o$), written $|w|$ (resp.\ $|o|$), equals the number of occurrences of labels in the sequence. 

For $i \in \mathbb{N}_{> 0}$, we define $X_i \colon \Path \to S$, $Y_i \colon \Path \to L$, $A_i \colon \Path \to A$, and $O_i \colon \Path \to \Obs_{\text{fin}}$ respectively as 
$X_i(\pi) = s_i$, $Y_i(\pi) = \ell_i$, $A_i(\pi) = a_i$, and $O_i(\pi) = (\ell_1, a_1) \cdots (\ell_{i-1}, a_{i-1}) \ell_{i}$, where $\pi = (\ell_1, s_1, a_1) (\ell_2, s_2, a_2) \cdots$.

\newcommand{\cyl}[1]{\mathit{cyl}(#1)}
Following the classical cylinder set construction~\cite[Ch10]{BaierK08-book}, we define the measurable space of paths $(\Path, \Sigma)$ where $\Sigma = \sigma(\{\cyl{w} \mid w \in \Path_{\text{fin}} \})$ is the smallest $\sigma$-algebra that contains all the \emph{cylinder sets} $\cyl{w} = w (A \times S \times L)^\omega$. 

To define a probability measure for MDPs, we use \emph{schedulers} (a.k.a., policies or strategies) to resolve the nondeterministic choices of actions that are taken at each step. 

A scheduler is a function $\sigma \colon \Path_{\text{fin}} \to \distr{A}$. Intuitively, a scheduler determines a distribution of actions to take, based on the \emph{history} of the current path. This notion of scheduler encompasses well-studied classes of schedulers such as memoryless, deterministic, and randomised~(\cf~\cite{BaierK08-book}). 
In this paper we distinguish between two types of schedulers, namely \emph{model-based} and \emph{observation-based} schedulers. A model-based scheduler chooses actions having complete knowledge of the \emph{history}. In contrast, an observation-based scheduler performs the choice based only on observable features of the history. 
\begin{definition}
A scheduler $\sigma$ is observation-based if for all $w, w' \in \Path_{\text{fin}}$ such that $|w| = |w'|$, $O(w) = O(w')$ implies $\sigma(w) = \sigma(w')$. 
\end{definition}

An MDP $\M$ and a scheduler $\sigma$ induce a probability space $(\Path, \Sigma, Pr^{\M}_\sigma)$ where $Pr^{\M}_\sigma$ denotes the (unique) probability measure such that for arbitrary $w =  (\ell_1, s_1,a_1) \cdots (\ell_{n-1}, s_{n-1},a_{n-1})(\ell_n, s_n) \in \Path_{\text{fin}}$, 
\begin{equation*}
Pr^{\N}_\sigma(\cyl{w}) = \textstyle \iota(\ell_1, s_1) \cdot \prod_{i = 1}^{n-1} \sigma(w_i)(a_i) \cdot
\tau_{a_i}(s_i)(\ell_{i+1}, s_{i+1}),
\end{equation*}
where $w_i = (\ell_1, s_1, a_1) \cdots (\ell_{i-1}, s_{i-1},a_{i-1}) (\ell_i, s_i)$ is the $i$-th prefix of $w$.

\section{Learning MPDs using Baum-Welch algorithm}\label{sec:passive}
In this section we present a variant of the Baum-Welch algorithm~\cite{Rabiner89} for learning an MDP $\M$ from a finite set of observation sequences $\bigo \subseteq \Obs_{\text{fin}}$.

As the Baum-Welch algorithm, also our method is a maximum likelihood approach: the transitions probabilities of $\M$ are estimated to maximise the likelihood 
$$L(\M,o) = Pr^\M[Y_{1:T} = \ell_1 \twodots \ell_T | A_{1:T-1} = a_1 \twodots a_{T-1}]$$
of an observed sequence $o = (\ell_1,a_1) \cdots (\ell_{T-1},a_{T-1}) \ell_T$. 
The maximum likelihood problem is solved using the expectation maximisation approach~\cite{Dempster77}. In this line, our algorithm starts with an initial model hypothesis $\hyp_0$ which is iteratively updated in a way that the likelihood is nondecreasing at each step, that is $L(\hyp_{n}) \leq L(\hyp_{n+1})$, until the likelihood difference between the current and the previous hypothesis goes below a  
fixed threshold $\epsilon$ (\emph{cf.}\ Figure~\ref{alg:mdpbw}).
\begin{figure}[t]
\begin{codebox}
\Procname{$\proc{Mdp-BW}(\bigo, \hyp_0)$}
\li $i \gets 0$
\li \Repeat 
\li $(\alpha, \beta) \gets \proc{Forward-Backward}(\hyp_i,\bigo)$
\li $\hyp_{i+1} \gets \proc{Update}(\hyp_i, \bigo, \alpha, \beta)$
\li $i \gets i + 1$
\li \Until $L(\hyp_{i},\bigo) - L(\hyp_{i-1},\bigo) \leq \epsilon$ 
\li \Return $\hyp_i$
\end{codebox}
\caption{Baum-Welch algorithm for MPDs} \label{alg:mdpbw}
\end{figure}

Next, we describe the update procedure. 
To ease the exposition, we fix the set of states $S$, labels $L$, and actions $A$ and we implicitly refer to the current hypothesis as the pair $\hyp = \langle \iota, \{\tau_a\}_{a \in A} \rangle$.
We define the forward and the backward functions $\alpha_o, \beta_o \colon S \times \{1\twodots T\} \to [0,1]$ for an observation sequence $o$ as
\begin{align*}
\alpha_o(s,t) &= Pr^\hyp[Y_{1:t} = \ell_1 \twodots \ell_t, X_t = s | A_{1:t-1} = a_1 \twodots a_{t-1}] \,\text{, and} \\
\beta_o(s,t) &= Pr^\hyp[Y_{t+1:T} = \ell_{t+1} \twodots \ell_{T} | X_t = s, A_{t:T-1} = a_t \twodots a_{T-1}] \,.
\end{align*}
These can be calculated using dynamic programming according to the following recurrences
\begin{align}
\alpha_o(s,t) &= \begin{cases}
	\iota(\ell_1, s) &\text{if $t = 1$}\\
	\displaystyle \sum_{s' \in S} \alpha(s',t-1) \, \tau_{a_{t-1}}(s')(\ell_t, s) &\text{if $1 \,{<} \,t \,{\leq}\,T$}
\end{cases} \label{eq:forward_passive}
\\
\beta_o(s,t) &= \begin{cases}
	1 &\text{if $t = T$} \\
	\displaystyle \sum_{s' \in S} \beta(s',t+1) \, \tau_{a_{t}}(s)(\ell_{t+1}, s') &\text{if $1 \,{\leq} \,t \,{<}\,T$}
\end{cases} \label{eq:backward_passive}
\end{align}


%
%
Next, we define $\gamma_o \colon S \times \{1, \twodots, T\} \to [0,1]$ and the action-indexed family of functions $\xi_o^a \colon S \times \{1, \twodots, T-1\} \times L \times S \to [0,1]$ for $a \in A$ as
\begin{align}
\gamma_o(s, t) &= Pr^\hyp[X_t = s | O_T = o] \, , \label{eq:gamma} \\
\xi^a_o(s,t)(\ell, s') &= Pr^\hyp[X_t = s, Y_{t+1} = \ell, X_{t+1} = s' | O_T = o] \,. \notag
\end{align}
The above are related to $\alpha_o$ and $\beta_o$ as follows
\begin{align*}
\gamma_o(s, t) &= \frac{\alpha_o(s,t) \cdot \beta_o(s,t)}{\sum_{s' \in S} \alpha_o(s',t) \cdot \beta_o(s',t) } \notag \\[1ex]
\xi^a_o(s,t)(\ell, s') 
	&=  1_{a_t}(t) 1_{\ell_t}(\ell) \; \frac{\alpha_o(s,t) \tau_a(s)(\ell,s') \beta_o(s',t+1)}{\sum_{u \in S}\alpha_o(u,t) \cdot \beta_o(u,t) }
\end{align*}
Given the current hypothesis $\hyp = \langle S, \iota, \{ \tau_a \}_{a \in A} \rangle$ of the model and a multiset $\bigo$ of i.i.d.\ observation sequences $o^1,\dots, o^R \in \Obs_{\text{fin}}$ where the $r$-th observation sequence is $o^r = \ell^r_1,a^r_1, \dots, \ell^r_{T_r-1},a^r_{T_r-1}, \ell^r_{T_r}$, the procedure $\proc{Update}(\hyp, \bigo, \alpha, \beta)$ updates $\iota$ and $\{ \tau_a \}_{a \in A}$ as follows
\begin{align*}
\iota(\ell,s) &= \frac{\sum_{r = 1}^R 1_{\ell^r_1}(\ell) \cdot \gamma_{o^r}(s,1)}{R} \\
\tau_a(s)(\ell,s') &= \frac{ \sum_{r = 1}^R \sum_{t = 1}^{T_r} \xi^a_{o^r}(s,t)(\ell, s') }{ \sum_{r = 1}^R \sum_{t = 1}^{T_r} 1_a(a^r_t)\cdot \gamma_{o^r}(s,t)} \,.
\end{align*}

\begin{remark}
Depending on the specific scheduler employed to sample the observations one may incur in the situation where $\sum_{r = 1}^R \sum_{t = 1}^{T_r} \gamma_{o^r}(s,t) = 0$, indicating that the state $s$ does not play a role in the observed dynamics. In this case the update procedure leaves the distributions $\{\tau_a(s)\}_{a \in A}$ unchanged.  
\end{remark}

The above described procedure is easily adapted to Markov chains, which are MDPs with a single action. Hereafter we use \mcbw\ to explicitly refer to such adaptation. 

\subsection{Experimental Results}\label{sec:passiveExperiments}
I this section we compare the quality of the models learned using \mcbw\ and \mdpbw\ respectively against the current state-of-the-art passive-learning tools for Markov chains and Markov decision processes, namely \textsc{Alergia}~\cite{CarrascoO94} and \textsc{IOAlergia}~\cite{MaoCJNLN16}.
Before we proceed, we briefly recall how \textsc{Alergia} and \textsc{IOAlergia} work. Both algorithms start from a maximal tree-shaped probabilistic automaton representing the training set $\bigo$, which is iteratively reduced by recursive merging operations among compatible states. Compatibility among states is determined based on the Hoeffding test parametric on a given \emph{confidence} value $\alpha \in (0,1)$. 

Remarkably, these approaches are very efficient and enjoy convergence properties. 
However, \textsc{IOAlergia} converges to the original (canonical) model $\M$ only if it is \emph{deterministic}, i.e., for all $s,s',s'' \in S$, $\ell \in L$, and $a \in A$, if $\tau_a(s)(\ell,s') > 0$ and $\tau_a(s)(\ell,s'') > 0$, then $s' = s''$. Hence each observation sequence is assumed to be emitted by a unique path. 

As a consequence, if the MDP under learning is not deterministic \textsc{IOAlergia} can only learn a deterministic approximation of the model which has often a larger state space. 

Due to the nature of the model construction, \textsc{Alergia} and \textsc{IOAlergia} do not require (nor explicitly allow) the user to choose the size of the learned model (i.e. the number of states) upfront. However, it can be tuned by choosing the input confidence value of $\alpha$.

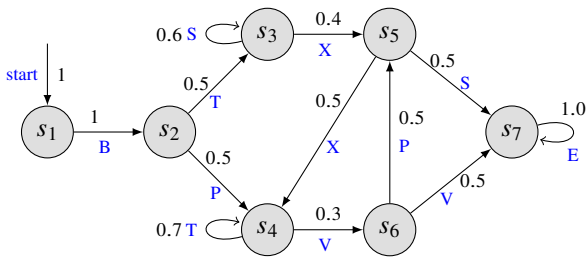
\begin{figure}[h!]
\centering
\begin{tikzpicture}[scale=1.3]

\node[vertex] (n1) at (0,0) {$s_1$};
	\node (init) at ($(n1)+(0,1)$) {};
\node[vertex] (n2) at ($(n1)+(1.25,0)$) {$s_2$};
\node[vertex] (n3) at ($(n2)+(1,1)$) {$s_3$};
\node[vertex] (n4) at ($(n2)+(1,-1)$) {$s_4$};
\node[vertex] (n5) at ($(n3)+(1.25,0)$) {$s_5$};
\node[vertex] (n6) at ($(n4)+(1.25,0)$) {$s_6$};
\node[vertex] (n7) at ($(n2)+(3.5,0)$) {$s_7$};

\draw[-latex] (init) --
	node[prob,right] {$1$} 
	node[label,left] {start} (n1);
\draw[-latex] (n1)--
	node[prob,above,pos=0.3] {$1$}
	node[label,below, pos=0.45] {B} (n2);
\draw[-latex] (n2)--
	node[prob,left,pos=0.5] {$0.5$}
	node[label,below, pos=0.45] {T} (n3);
\draw[-latex] (n2)--
	node[prob,above=0.1,pos=0.5] {$0.5$}
	node[label,below, pos=0.45] {P} (n4);
\draw[-latex] (n3)--
	node[prob,above,pos=0.5] {$0.4$}
	node[label,below, pos=0.45] {X} (n5);
\draw[-latex] (n4)--
	node[prob,above,pos=0.5] {$0.3$}
	node[label, below, pos=0.45] {V} (n6);
\draw[-latex] (n5)--
	node[prob,above,pos=0.4] {$0.5$}
	node[label,right, pos=0.5] {S} (n7);
\draw[-latex] (n6)--
	node[prob,right,pos=0.5] {$0.5$}
	node[label,below, pos=0.45] {V} (n7);
\draw[-latex] (n5)--
	node[prob,above=0.2,pos=0.5] {$0.5$}
	node[label,below=0.1, pos=0.45] {X} (n4);
\draw[-latex] (n6)--
	node[prob,right,pos=0.6] {$0.5$}
	node[label,right, pos=0.4] {P} (n5);
\draw[-latex] (n3) edge[loop left] 
	node[prob,left=0.2] {$0.6$} 
	node[label,left] {S} (n3);
\draw[-latex] (n4) edge[loop left] 
	node[prob,left=0.2] {$0.7$} 
	node[label,left] {T} (n4);
\draw[-latex] (n7) edge[loop right] 
	node[prob,above=0.1] {$1.0$} 
	node[label,below=0.1] {E} (n7);

\end{tikzpicture}
\caption{The REBER grammar from~\cite{Reber67}}
\label{figure:reber}
\end{figure}

\paragraph*{\mcbw\ vs. \textsc{Alergia}} For experimental comparison between \mcbw\ and \textsc{Alergia}, we fixed a \emph{training set} $\bigo$ and a \emph{test set} $\mathcal{T}$ respectively consisting of $10^4$ and $10^5$  observation sequences of length $5$ generated by the chain in Figure~\ref{figure:reber}. The size of the test set is $10$ times bigger than that of the training set because we are interested in measuring to what extent the learning procedures are able to generalise w.r.t.\ a relatively small training set. First we have run \mcbw\ starting from a random initial hypothesis with $n = 7 \twodots 15$ states, then we have run \textsc{Alergia} with an input value of $\alpha$ chosen to match the size of the learned model to $n$.

\begin{table*}[t]
\small \setlength{\tabcolsep}{1ex}
\centering
\subfloat[Comparison of Alergia and \mcbw\ on the \textsc{Reber} grammar from~\cite{Reber67}.]{
\begin{tikzpicture}[baseline=(table1.center)]
\node (table1) at (0,0) {
\begin{tabular}[t]{|c|c|c|c|c|c|c|c|}
\hline
 \multirow{2}{*}{$|S|$} &
 \multicolumn{4}{c|}{\textsc{Alergia}} &
 \multicolumn{3}{c|}{\textsc{Mc-BW}}
 \\ \cline{2-8}
 & $\alpha$ & $\ln L$ on $\bigo$ & $\ln L$ on $\T$ & KL div. & $\ln L$ on $\bigo$ & $\ln L$ on $\T$ & KL div. \\ \hline\hline
 7 & 2.09e-201 & $-3.968$ & $-4.163$ & $1.256$ & $-2.597$ & $-2.66$ & $0.086$ \\ \hline 
 8 & 7.28e-160 & $-3.836$ & $-4.239$ & $1.025$ & $-2.595$ & \textbf{-2.651} & $0.086$ \\ \hline 
 9 & 2.93e-100 & $-3.257$ & $-3.432$ & $0.607$ & $-2.597$ & $-2.659$ & $0.086$ \\ \hline 
 10 & 7.14e-104 & $-2.993$ & $-3.133$ & $0.376$ & $-2.587$ & $-2.654$ & $0.095$ \\ \hline 
 11 & 5.66e-75 & $-3.076$ & $-3.231$ & $0.29$ & $-2.693$ & $-2.808$ & \textbf{0.001} \\ \hline 
 12 & 2.87e-44 & $-2.701$ & $-2.804$ & $0.002$ & $-2.699$ & $-2.807$ & \textbf{0.001} \\ \hline 
 13 & 0.01 & $-2.701$ & $-2.803$ & $0.002$ & $-2.54$ & $-2.72$ & $0.155$ \\ \hline 
 14 & 0.5 & $-2.693$ & \textbf{-2.8} & \textbf{0.001} & $-2.586$ & $-2.657$ & $0.095$ \\ \hline 
 15 & 0.9 & $-2.694$ & $-2.808$ & $0.001$ & $-2.533$ & $-2.723$ & $0.161$ \\ \hline 
\end{tabular}};
\end{tikzpicture}
\label{tab:MCexperiments}
}
\subfloat[Comparison of \textsc{IOAlergia} and \mdpbw\ on an adaptation of the Grid World model from~\cite{TapplerA0EL19}.]{%
\begin{tikzpicture}[baseline=(table2.center)]
\node (table2) at (0,0) {
 \begin{tabular}{|c|c|c|c|}
    \hline 
     & $\ln L$ on $\bigo$ & $\ln L$ on $\T$ & KL div. \\ \hline
    True model & $-4.171$ & $-4.262$ & $0$ \\ \hline
    \mdpbw & $-4.899$ & $-4.989$ & $0.333$ \\ \hline
    \textsc{IOAlergia} & $-13.83$ & $-$ & $-$ \\ \hline 
 \end{tabular}}; 
 \end{tikzpicture}
    \label{tab:MDP-BWvsIOAlergia}
    }
    \caption{Comparative analysis of the Baum-Welch algorithm vs Alergia} 
    \label{tab:PassiveExperiments}
\end{table*}

\begin{figure}[th]
\centering
\includegraphics[scale=0.4]{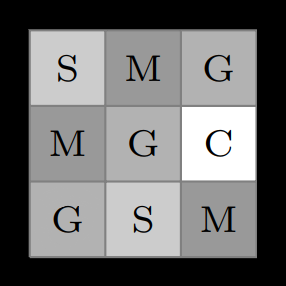}
\caption{The Small Grid World Model.}
\label{figure:small_grid}
\end{figure}

Table~\ref{tab:MCexperiments} summarises the results of our experiments 
in terms of the quality of the learned models. The values reported in the table correspond to the loglikelihood of $\bigo$ (resp.\ $\T$) divided by $|\bigo|$ (resp.\ $|\T|$) and the Kullback-Leibler divergence relative to $\T$.
We can see that \mcbw\ achieves better quality performace with fewer states compared with \textsc{Alergia}. 
Interestingly, we observe an increased size of the model does not necessarily correspond to a quality improvement.  This phenomenon may have two plausible explanations: 
\begin{enumerate*}[label=(\roman*)]
\item having too many states leads the learning procedure to overfit the training set; 
\item or only a portion of the model gets updated by the procedure, while the remaining portion of the model is left almost identical to the starting hypothesis. 
\end{enumerate*}
\paragraph*{\mdpbw\ vs. \textsc{IOAlergia}} 
By using the same methodology, we compared \mdpbw\ against \textsc{IOAlergia}\cite{MaoCJNLN16}.\\
Here the model we are learning is a smaller variant of the grid world introduced in \cite{TapplerA0EL19} (\cf\ Figure~\ref{figure:small_grid}). A robot is moving in this grid, starting from the middle cell. The actions are the four directions ---nord, east, south, and west--- and the observed labels represent different terrains. Depending on target terrain the robot may slip and change direction, e.g. move south west instead of south. By construction, the model is a deterministic MDP thus, in the big sample limit, \textsc{IOAlergia} can learn it. 

For the comparison, we used a \emph{training set} $\bigo$ and a \emph{test set} $\T$ consisting respectively of $10^3$ and $10^2$ sequences of $10$ length. With $\alpha=0.05$, \textsc{IOAlergia} produced a model with $10$ states. We then run \mdpbw\ staring from a randomly generated initial hypothesis with $9$ states. Table~\ref{tab:MDP-BWvsIOAlergia} summarises the results of the comparison. 
On the training set, the model learned by \textsc{IOAlergia} scores lower log-likelihood value than the model learned by \mdpbw. Notably, the test set had a number of observations that could not be generated by the model produced with \textsc{IOAlergia}. In contrast, the MDP learned with \mdpbw\ was able to generalise better from the training set, achieving a log-likelihood value on $\T$ comparably similar to the one measured on original grid-world model.  
This results show us that for small training sets, \mdpbw\ seems to attain more accurate models than \textsc{IOAlergia}, which requires big training sets to achieve good results. 

However, the price of the accuracy of \mdpbw\ is payed in terms of efficiency: in all experiments \textsc{IOAlergia} run orders of magnitude faster than \mdpbw. This is not surprising, because \textsc{IOAlergia} has a run-time complexity that grow linearly in the size of the data set.

\section{Active Learning of Markov Decision Processes}\label{sec:active}
\newcommand{\graphheight}{28ex}
\begin{figure*}[h!]
\centering
%
\subfloat[Street crossing model: log-likelihood graphs relative to a test set of of $200$ sequences of fixed length $12$.]{\includegraphics[height=\graphheight]{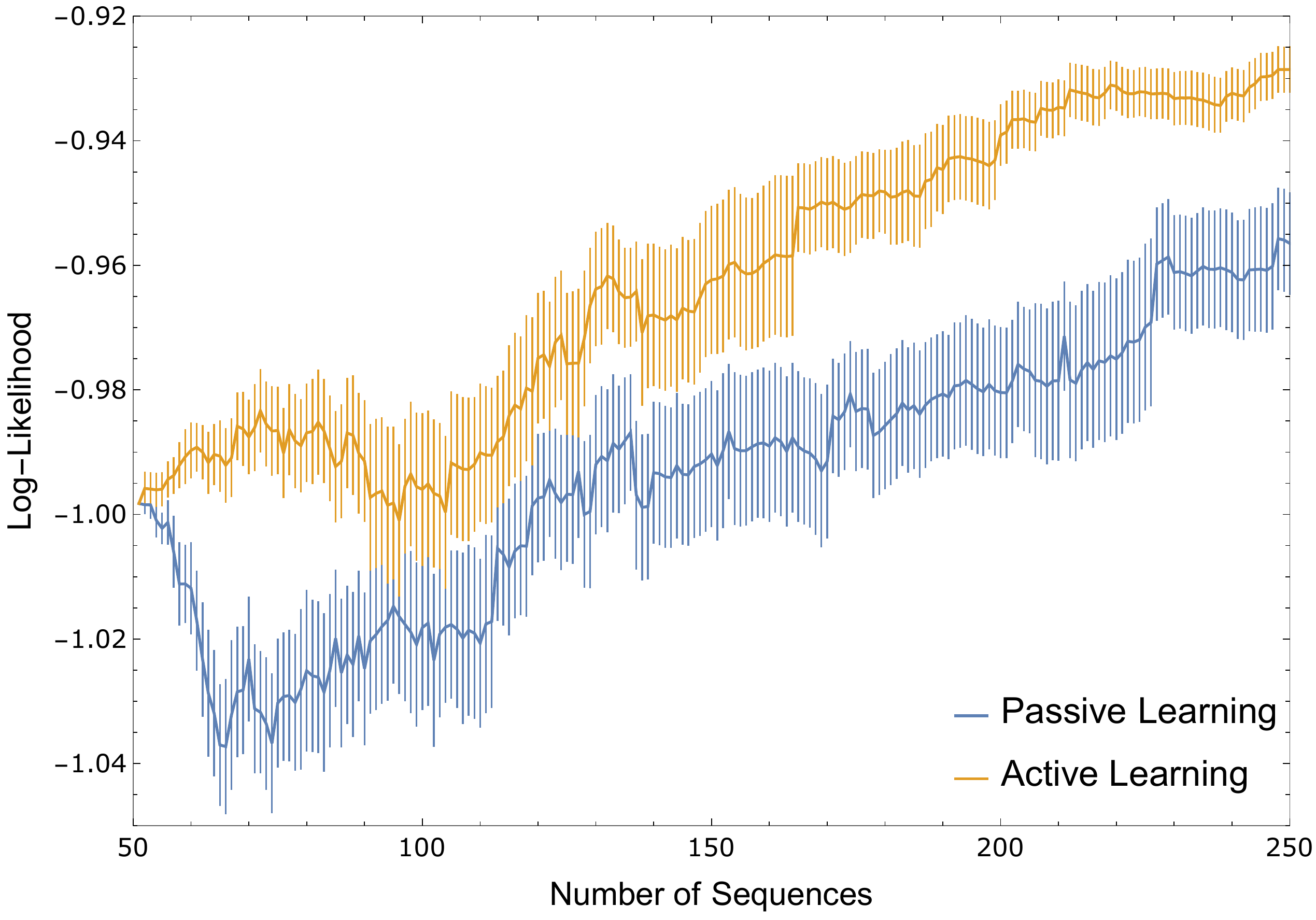}
\label{fig:test-active-passive:big-street}}
\hfil
\subfloat[Small grid world model: log-likelihood graphs relative a test set of of $200$ sequences of length $T \sim \text{Geo}(0.8)$.]{
\includegraphics[height=\graphheight]{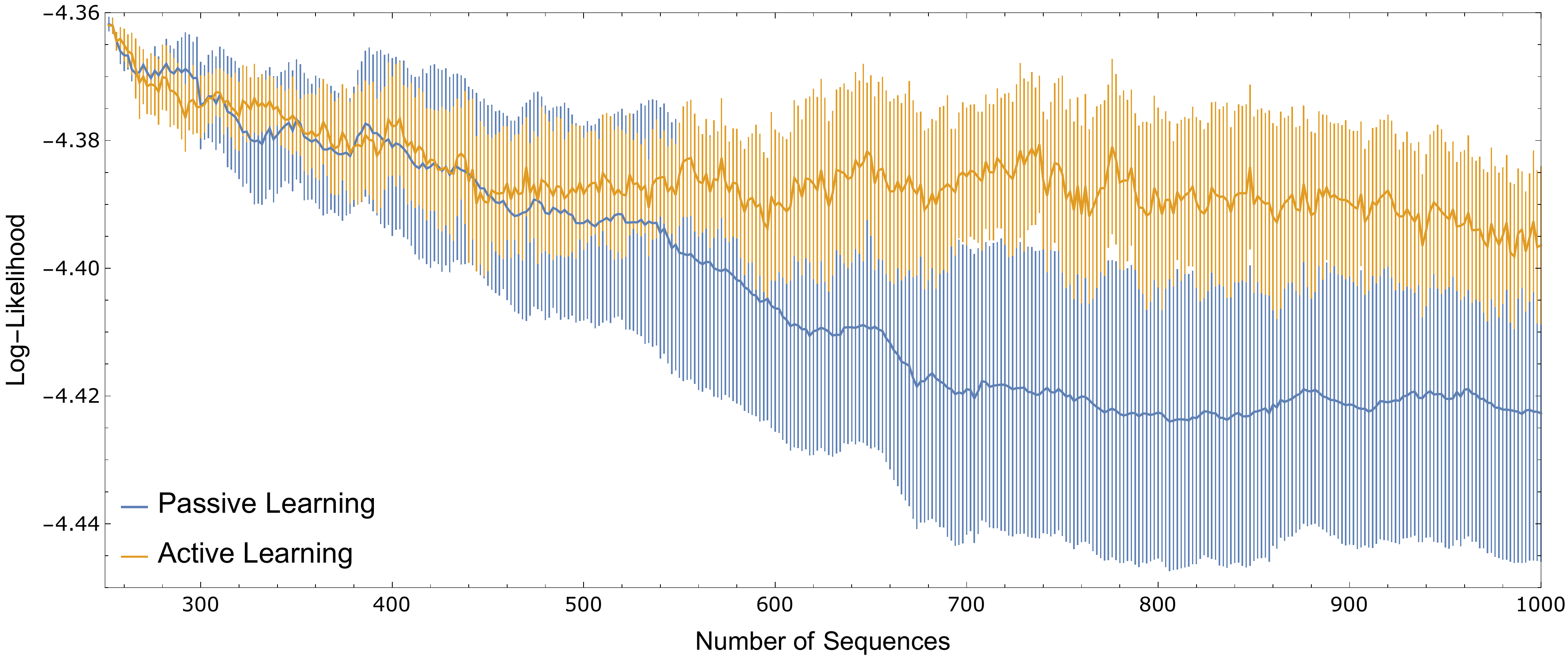}
\label{fig:test-active-passive:gridworld}}
\caption{Comparison between the passive learning and active learning procedures based on the \mdpbw\ algorithm.}
\label{fig:active-passive_graphs}
\end{figure*}

The \textsc{Mdp-BW} algorithm is a passive learning method: it assumes no interaction with the system, which has to be learned from a fixed set of observations. 
In situations where one can \emph{actively query} the system to collect training data, one can think of employing querying strategies to produce new examples that are most informative w.r.t.\ the systems nondeterministic behaviour. In this way, one can learn qualitatively better models compared to the passive learning approach while collecting a considerably smaller amount of observations.

Let $\hyp = \langle S,A, \iota, \{ \tau_a \}_{a \in A} \rangle$ and $\bigo = \{o^1,\dots, o^R\}$ be respectively the current hypothesis and the current training set. The active learning procedure iteratively updates $\hyp$ and $\bigo$ by performing the following steps:
\begin{enumerate}
\item devise an observation-based scheduler from $\bigo$ and $\hyp$;
\item sample new observation sequences using the above mentioned scheduler, adding them to $\bigo$; and
\item update $\hyp$ based on the new data using \mdpbw.   
\end{enumerate}
These steps are repeated until a given sampling budget has been exceeded or no further scrutiny of the system is deemed necessary. Hereafter, we detail how each step is implemented.

We start by computing the matrix $M = (m_{sa})_{s \in S, a \in A}$ where $m_{sa}$ is the expected number of times the action $a$ has been chosen from $s$, that is computed as follows
\begin{equation}
m_{sa} = \textstyle \sum_{r = 1}^{R} \sum_{t = 1}^{|o^r|} 1_{a}(a_t^r) \, \gamma_{o^r}(s,t) \,, 
\label{eq:matrixM}
\end{equation}
then, we define the memoryless scheduler $\sigma_M \colon S \to \distr{A}$ as
\begin{equation}
\textstyle \sigma_M(s)(a) = 1 - (m_{sa} / \sum_{a' \in A} m_{sa'}) \,.
\label{eq:oppositescheduler}
\end{equation}
Intuitively, given the system is in state $s \in S$, the above scheduler chooses an action $a \in A$ with a probability that is opposite to that observed in $\bigo$. Since the current state of the system is hidden, when sampling we use a belief state instead. This corresponds to employ the observation-based scheduler $\sigma_M^* \colon \Obs_{\text{fin}} \to \distr{A}$ defined as follows.
For an observation $o = (\ell_1,a_1) \cdots (\ell_{t-1},a_{t-1}) \ell_t \in \Obs_{\text{fin}}$ and an action $a \in A$, 
\begin{align}
\sigma_M^*(o)(a) &= \textstyle \sum_{s \in S }\Pb^\hyp[X_t = s | O_t = o] \cdot \sigma_M(s)(a) \notag\\
&= \textstyle \sum_{s \in S} \gamma_o(s,t) \, \sigma_M(s)(a) \,. \label{eq:obssched}
\end{align}
Intuitively, the above scheduler works as follows. Having observed $o$, we believe system is in state $s \in S$ with probability $\Pb^\hyp[X_t = s | O_t = o]$; consequently, $\sigma^*_M$ chooses the action $a \in A$ with probability $\sigma_M(s)(a)$.

The algorithm in Fig.~\ref{alg:active-sampling} describes how we actively sample an observation sequence of length $T \in \mathbb{N}$ emitted by a partially observable MDP $\M$ by using the scheduler $\sigma_M^*$ of Eq.~\eqref{eq:obssched}.
\begin{figure}[h!]
\begin{codebox}
\Procname{$\proc{ActiveSampling}(\M,\hyp = \langle S, \iota, \{\tau_a\}_{a \in A} \rangle , \bigo , T \in \mathbb{N})$}
\li Initialise $M = (m_{sa})_{s \in S, a \in A}$ as Eq.~\eqref{eq:matrixM}
\li $\ell_1 = \proc{Init}(\M)$ \Comment{initialise the system}
\li \For \textbf{each} $s \in S$ \Do \label{lin:initbelief1}
\li $\alpha(s,1) = \iota(\ell_1,s)$
\End \label{lin:initbelief2}
\li \For $t = 1$ \To $T-1$ \Do
\li Sample $a_{t} \in A$ according to $\sum_{s \in S} \frac{\alpha(s,t)}{\sum_{s' \in S} \alpha(s',t)} \sigma_M(s)$ \label{lin:pickaction}
\li $\ell_{t+1} = \proc{Observe-Label}(\M, a_t)$ \label{lin:obsnext}
\li \For \textbf{each} $s \in S$ \Do \label{lin:updatebelief1}
\li $m_{s a_t} = m_{s a_t} +  \alpha(s,t) / \sum_{s' \in S} \alpha(s',t)$
\li $\alpha(s,t+1) = \sum_{s' \in S} \tau_{a_t}(s')(\ell_{t+1},s) \cdot \alpha(s',t)$
\End \label{lin:updatebelief2}
\End
\li \Comment{Return the entire observation sequence}
\li \Return $(\ell_1, a_1)\cdots(\ell_{T-1},a_{T-1}) \ell_T$
\end{codebox}
\caption{Active Sampling Strategy}
\label{alg:active-sampling}
\end{figure}

\proc{ActiveSampling} keeps track and updates at each step the matrix $M$ and the current forward distribution $\alpha(\cdot, t) \in \distr{S}$. These are respectively used to compute the current belief state $\gamma(\cdot, t) \in \distr{S}$ (\cf\ Eq.~\eqref{eq:gamma}) and the memoryless scheduler $\sigma_M$ (\cf\ Eq.~\eqref{eq:oppositescheduler}), which are used in line~\ref{lin:pickaction}.  
After observing the an initial label $\ell_1$ from the system $\M$, the initial forward distribution $\alpha(\cdot, 1)$ is computed (lines~\ref{lin:initbelief1}--\ref{lin:initbelief2}). Then, for each time-step $t$ from $1$ to $T - 1$, an action $a_t \in A$ is sampled according to $\sigma_M^*$, and used to observe the next label $\ell_{t+1}$ emitted by $\M$ (line~\ref{lin:obsnext}). The forward distribution $\alpha(\cdot, t + 1)$ and the matrix $M$ are then updated (line~\ref{lin:updatebelief1}--\ref{lin:updatebelief2}) before moving to the next time-step.
The update of the forward probabilities follows Eq.~\eqref{eq:forward_passive}, while the update of the column vector $M_{a_t}$ follows Eq.~\eqref{eq:matrixM}.

\subsection{Experimental Results}\label{sec:activeExperiments}
\begin{figure}[t!]
\centering
\begin{tikzpicture}[scale=1.3]

\node[vertex] (n1) at (0,0) {$s_1$};
	\node (init) at ($(n1)+(-0.2,+0.8)$) {};
	\node[bullet] (n1b) at ($(n1)+(-0.2,-0.8)$) {};
	\node[bullet] (n1bb) at ($(n1)+(1,-0.1)$) {};
\node[vertex] (n2) at ($(n1)+(5,0)$) {$s_2$};
	\node[bullet] (n2b) at ($(n2)+(0.2,-0.8)$) {};
	\node[bullet] (n2bb) at ($(n2)+(-1,0.2)$) {};
\node[vertex] (n3) at ($($(n1)!.5!(n2)$)+(0,-1)$) {$s_3$};
\node[vertex] (hit) at ($(n1)+(1,-2)$) {hit};
\node[vertex] (ok) at ($(n2)+(-1,-2)$) {ok};

\draw[-] (n1)--node[action,left] {stay} (n1b); 
\draw[-] (n1)--node[action,above=-0.05, pos=0.6] {move} (n1bb); 
\draw[-] (n2)--node[action,right] {stay} (n2b); 
\draw[-] (n2)--node[action,above] {move} (n2bb); 

\draw[-latex] (init) --
	node[prob,right] {$1$} 
	node[label,left] {start} (n1);
\draw[-latex] (n1b)--
	node[prob,below,pos=0.3] {$p$}
	node[label,below=-0.05, pos=0.45] {left} (n3);
\draw[-latex] (n1b) edge[bend right]
	node[prob,pos=0.3, right] {$1-p$}
	node[label,pos=0.8, right] {right} (n1);
\draw[-latex] (n1bb) edge
	node[prob,pos=0.4, left] {$1-p$}
	node[label,pos=0.7, left] {right} (n3);
\draw[-latex] (n1bb) edge[bend right=10]
	node[prob,pos=0.2, above=-0.05] {$p$}
	node[label,pos=0.3, above] {left} (n2);
		
\draw[-latex] (n2b)--
	node[prob,below,pos=0.3] {$p$}
	node[label,below=-0.05, pos=0.45] {right} (n3);
\draw[-latex] (n2b) edge[bend left]
	node[prob,pos=0.3, left] {$1-p$}
	node[label,pos=0.8, left] {left} (n2);
\draw[-latex] (n2bb) edge
	node[prob,pos=0.6, right] {$1-p$}
	node[label,pos=0.8, right] {left} (n3);
\draw[-latex] (n2bb) edge[bend right=15]
	node[prob,pos=0.2, below=-0.05] {$p$}
	node[label,pos=0.3, below=-0.05] {right} (n1);
\draw[-latex] (n3)--
	node[action,above] {move} 
	node[prob,left] {$1$} 
	node[label,right] {bump} (hit);
	
\draw[-latex] (n3)--
	node[action,above] {stay} 
	node[prob,left] {$1$} 
	node[label,right] {avoid} (ok);

\draw[-latex] (ok) edge[loop right] 
	node[action,above] {stay}
	node[prob,right] {$1$} 
	node[label,right=0.2] {avoid} (ok);
\draw[-latex] (ok) edge[loop below] 
	node[action,right,pos=0.1] {move}
	node[prob,right] {$1$} 
	node[label,right=0.2] {avoid} (ok);
	
\draw[-latex] (hit) edge[loop left] 
	node[action,above] {stay}
	node[prob,left] {$1$} 
	node[label,left=0.2] {bump} (hit);
\draw[-latex] (hit) edge[loop below] 
	node[action,left,pos=0.9] {move}
	node[prob,left] {$1$} 
	node[label,left=0.2] {bump} (hit); 
\end{tikzpicture}
\caption{The Street crossing model} \label{figure:big-street}
\end{figure}
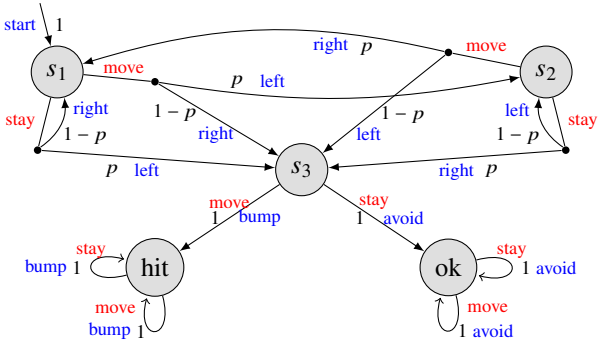
In this section we present an empirical analysis of the active sampling strategy. We will use two case study models: the small grid world model from previous section (see Fig.~\ref{figure:small_grid}), and the street crossing model (depicted in Fig.~\ref{figure:big-street}). 
The former model represents an agent trying to avoid a stranger bumping into her. Here she can choose among two actions: \emph{stay} on the current side of the sidewalk or \emph{move} to the other side. The agent and the stranger make their move independently at the same time; in particular, when the two are not in front each other the stranger, proceeds forward. After performing the action, the agent observes if the stranger is on the \emph{left} or the \emph{right} side of the street. 
If the two end up in the same side they \emph{bump} into each other, otherwise they \emph{avoid} each other. The stranger changes side with probability $p \in (0,1)$.

We compare the active procedure against the passive one and show how the learning accuracy of the former compares to the latter with the size of the training set. 
The experiments have been performed as follows. Starting from the same initial hypothesis ---learned with \mdpbw\ from a small data set--- we incrementally grew the data set bigger respectively using the active sampling strategy and a sampling strategy based on a memoryless uniformly distributed selection of actions.   
For the street crossing model the initial hypothesis was learned from a data set of 50 sequences of length 12; then we performed 200 active learning iterations. Fig.~\ref{fig:test-active-passive:big-street} shows the graph of the mean log-likelihood paired with standard error bars measured from a number of re-run of the experiment relative to test set of $200$ sequences each of length $12$. 

For the small grid world model the initial hypothesis was learned from 250 observation sequences of length $T$ distributed according to a geometric distribution with success probability $p = 0.8$, that is $T \sim \text{Geo}(0.8)$; then we performed $750$ active learning iterations by sampling new observations of length $T \sim \text{Geo}(0.8)$. Analogously to the first case study, the results of this experiment are summarised in Fig.~\ref{fig:test-active-passive:gridworld}. The graph shows that the passive learning approach has a more pronounced tendency to overfit the data set than the active learning approach. 

Overall, the graphs in Fig.~\ref{fig:active-passive_graphs} show that the active learning approach provides better approximations than the passive approach. Another interpretation is that the proposed active learning is able to obtain the same level of accuracy than the passive learning approach with a smaller data set. Notably, the graphs show also that the standard error for the active learning method is smaller than the one measured for the passive learning approach. This indicates that our active learning approach is more stable than the passive approach.
\begin{figure}[t]
\centering
\includegraphics[scale=0.4]{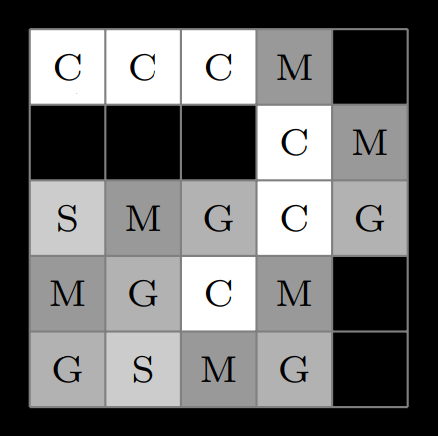}
\caption{The Grid World Model from~\cite{TapplerA0EL19}.}
\label{figure:grid}
\end{figure}

\newcommand{\LstarMdp}{${L_\textsc{mdp}^*}$}
\begin{table}[t]
\small \setlength{\tabcolsep}{0.5ex}
\centering
\begin{tabular}{r|c|c|c|c|}
 & true & \LstarMdp & \textsc{IOAlergia} & \textsc{A-\mdpbw} \\\hline
overall \# of labels     & - & $3101959$ & $3103607$ & $23781$\\\hline
\# of observation traces      & - & $391530$  & $387746$ & $1200$\\\hline
$|S|$ (\# of states)    & 35 &  $35$ & $21$ & $19$\\\hline \hline
bismilarity distance $\delta_{0.9}$ & 0 & $0.144$ & $0.524$ & $0.364$\\\hline \hline
$\mathbb{P}_{\max}(F^{<12} (\mathrm{goal}))$ & $0.962$ & $0.965$  & $0.230$ & $0.978$\\\hline
$\mathbb{P}_{\max}(\lnot \mathrm{G}\ U^{\leq14} (\mathrm{goal}))$ & $0.65$ & $0.646$  & $0.158$ & $0.466$\\\hline
$\mathbb{P}_{\max}(\lnot \mathrm{S}\ U^{\leq16} (\mathrm{goal}))$ & $0.691$ &  $0.676$ & $0.180$ & $0.806$\\
\end{tabular}
 
\caption{Results for learning the grid world model.}
 \label{tab:results_gridworld_1}
\end{table}
\paragraph*{Active \mdpbw\ vs \LstarMdp} We conclude the experiment section by comparing our active learning method against the \LstarMdp\ algorithm~\cite{TapplerA0EL19} for learning deterministic MDPs. We recall that \LstarMdp\ actively refines its current hypothesis as long as the teacher can provide new counterexamples. The implementation of the teacher in the \LstarMdp\ algorithm is done both by checking the conformance and the structure of the hypothesis w.r.t\ the data set. 

For the comparison we replicated the same experiment performed in~\cite{TapplerA0EL19} for comparing \textsc{IOAlergia} with \LstarMdp\ when learning the grid world model depicted in Fig.~\ref{figure:grid}. 

Our model was learned using the active learning approach starting from a (deterministic) initial model with $19$ states, learned from a small dataset of $200$ sequences. The length $T$ of each sampled sequence is distributed according to a geometric distribution shifted by $10$ with success probability $p = 0.9$, that is, $T \sim 10 + \text{Geo}(0.9)$\footnote{Specifically, $P(T = 10+ k) = (1-p)^{k-1} p$ for $k \in \mathbb{N}_{>0}$.}. At each active learning iteration we sampled two new sequences, and we stopped after collecting $1200$ observation traces. 
Table~\ref{tab:results_gridworld_1} shows the results of the experiment. 
As done in~\cite{TapplerA0EL19} we compared the models with respect to the bisimilarity distance\footnote{To compute the distance, we used the MDPDist library~\cite{BacciBLM13} adapted to labelled MDPs.} with discount factor $\lambda = 0.9$: the model learned with our active learning approach, scores slightly better than \textsc{IOAlergia} but worse than \LstarMdp. Nevertheless, the results of the three model-checking queries performed on our model are close to the true one: the absolute error from the true values is bounded by $0.184$. Overall, \LstarMdp\ scores better than our active learning approach. This is due to a number of reasons: 
\begin{enumerate*}[label=(\roman*)]
\item the learned model is smaller than the canonical true model and
\item it was learned from a significantly smaller data set; finally,
\item the active learning approach is not sensitive to structural counterexamples as the \LstarMdp\ algorithm is.
\end{enumerate*}
Indeed, when the algorithm encounters a new observation which has probability zero of being generated by the current hypothesis, also the next hypothesis won't be able to generate it. This aspect in particular needs particular attention when learning deterministic models or in general when some observation traces can be emitted only by a single path in the hypothesis model.

\section{Conclusions and Future Work}\label{sec:conclusion}
In this paper we revisited the classic Baum-Welch algorithm for learning models parameters of \emph{nondeterministic} MDPs and Markov chains from a set of observations. Compared with state-of-the-art (passive) learning algorithms like \textsc{Alergia} and \textsc{IOAlergia}, the \mdpbw\ procedure has a higher run-time complexity. However, experiments show that \mdpbw\ is able to learn models that reflect more accurately the behaviours of the observed system. This aspect is more pronounced when learning MDPs from a relatively small set of observations. 

Learning model parameters for MDPs typically requires large data sets, especially when the system under learning exhibits a high degree of nondeterminism. To cope with this issue, we proposed a model-based active learning sampling strategy which has three main advantages: 
\begin{enumerate*}[label=(\alph*)]
\item it is simple to implement and can be seamlessly integrated into small low power embedded systems;  
\item it does not introduce additional overhead with respect to the model update procedure; 
\item it collects a diverse and well-spread variety of observations, that better represent the nondeterministic behaviours of the system under learning.
\end{enumerate*}
Experimental results show that the active procedure strategy outperforms the corresponding passive learning variant in terms of accuracy relative to the size of the data set. This makes our active learning procedure an effective solution when one has the possibility to have limited amount of interactions with the system under learning.

A weakness of our active learning procedure is the fact that is it not sensitive to structural counterexamples. As future work we intend address this issue. 

Another interesting research direction consists in generalising the active learning procedure for learning model parameters of stochastic two-player games, allowing one to learn systems that operate in an unknown (adversarial) environment by actively interacting with both players.

\bibliographystyle{IEEEtran}
\bibliography{bibliography}

\end{document}